\documentclass[letterpaper, 10 pt, conference]{ieeeconf}  

\IEEEoverridecommandlockouts                              

\title{\LARGE \bf
GaussianRoom: Improving 3D Gaussian Splatting with SDF Guidance and Monocular Cues for Indoor Scene Reconstruction
}

\author{
    Haodong Xiang\dag, \ 
    Xinghui Li\dag, \ 
    Kai Cheng\dag, \ 
    Xiansong Lai, \ 
    Wanting Zhang, \ 
    Zhichao Liao, \ \\ 
    Long Zeng*, \  
    Xueping Liu* \ \\
    \vspace{-0.8cm}
\thanks{\dag\ Equal contribution }
\thanks{* Corresponding author. Emails:\{liuxp, zenglong\}@sz.tsinghua.edu.cn}
\thanks{Haodong Xiang, Xinghui Li, Xiansong Lai, Wanting Zhang, Zhichao Liao, Long Zeng, and Xueping Liu are with Tsinghua Shenzhen International Graduate School, Tsinghua University, Shenzhen, China.}
\thanks{Kai Cheng is with University of Science and Technology of China, Hefei, China.}
}

\usepackage{graphicx}
\usepackage{amsmath}
\usepackage{amssymb}
\usepackage{booktabs}
\usepackage{xcolor} 
\usepackage{colortbl}

\begin{document}

\maketitle
\thispagestyle{empty}
\pagestyle{empty}

\begin{abstract}

Embodied intelligence requires precise reconstruction and rendering to simulate large-scale real-world data. Although 3D Gaussian Splatting (3DGS) has recently demonstrated high-quality results with real-time performance, it still faces challenges in indoor scenes with large, textureless regions, resulting in incomplete and noisy reconstructions due to poor point cloud initialization and underconstrained optimization.
Inspired by the continuity of signed distance field (SDF), which naturally has advantages in modeling surfaces, we propose a unified optimization framework that integrates neural signed distance fields (SDFs) with 3DGS for accurate geometry reconstruction and real-time rendering.
This framework incorporates a neural SDF field to guide the densification and pruning of Gaussians, enabling Gaussians to model scenes accurately even with poor initialized point clouds. 
Simultaneously, the geometry represented by Gaussians improves the efficiency of the SDF field by piloting its point sampling. Additionally, we introduce two regularization terms based on normal and edge priors to resolve geometric ambiguities in textureless areas and enhance detail accuracy. Extensive experiments in ScanNet and ScanNet++ show that our method achieves state-of-the-art performance in both surface reconstruction and novel view synthesis. 
Project page: \textcolor{blue}{https://xhd0612.github.io/GaussianRoom.github.io/} 

\end{abstract}

\section{INTRODUCTION}

3D reconstruction transforms 2D images into detailed 3D models, enabling high-fidelity rendering of real-world scenes.
It plays crucial roles in visual simulation for embodied intelligence, bridging the gap between simulated environments and real-world applications~\cite{li20223d, jing2024two, chen2024fusednet, tang2024mobile}.

One typical scenario for reconstruction is indoor scenes, which is characterized by large textureless areas. 
MVS-based methods \cite{schonberger2016pixelwise, yao2018mvsnet} often produce incomplete or geometrically incorrect reconstruction results, primarily due to the geometric ambiguities introduced by the textureless areas.
Recently, neural-radiance-filed-based methods \cite{wang2021neus, wang2022neuris, yu2022monosdf} that model the scenes using signed distance field (SDF) have achieved complete and accurate mesh reconstruction in indoor scenes, benefiting from the continuity of neural SDFs and the introduction of monocular geometric priors \cite{yu2022monosdf}.
However, they suffer from long optimization times due to dense ray sampling in volume rendering.
Fortunately, 3D Gaussian Splatting (3DGS) \cite{kerbl20233dgs} accelerates the optimization and rendering speed of neural rendering by its differentiable rasterization technique, which also provides a new possibility in 3D scene reconstruction. Despite its impressive rendering efficiency, 3DGS often gets noisy,  and incomplete reconstruction results in indoor scenes. This is primarily due to the poor initialization of the SfM point cloud in textureless regions and the under-constrained densification and optimization of Gaussians.

Considering both the advantages of neural SDF in modeling surfaces and the efficiency of 3DGS, we introduce a novel approach named \textbf{GaussianRoom}, 
which incorporate neural SDF within 3DGS to improve geometry reconstruction in indoor scenes while preserving rendering efficiency.
We design a jointly optimizing strategy to enable 3DGS and the neural SDF to facilitate each other. 

First, we propose a \textbf{SDF-guided primitive distribution strategy}, which utilizes surfaces represented by SDF to guide the densification and pruning of Gaussian. For surface areas lacking initial Gaussians, we deploy new Gaussians using the \textbf{SDF-guided Global Densification strategy}. For the existing Gaussians, we perform \textbf{SDF-guided Densification and Pruning} based on their positions relative to the scene surface. At the same time, the Gaussians are used to guide point sampling of the neural SDF along the rays, which reduces invalid sampling in free space and further improves the efficiency of optimization.
Secondly, considering that rendering loss is insufficient to constrain textureless areas, we additionally introduce \textbf{monocular normal priors} in both Gaussian and neural SDF fields to regularize the geometry of textureless areas.
Furthermore, to improve the details of indoor scenes, we increase the sampling of pixels covering fine structures for training based on the edge prior.


\begin{figure*}[t]
    \centering
    \includegraphics[width=0.8\textwidth]{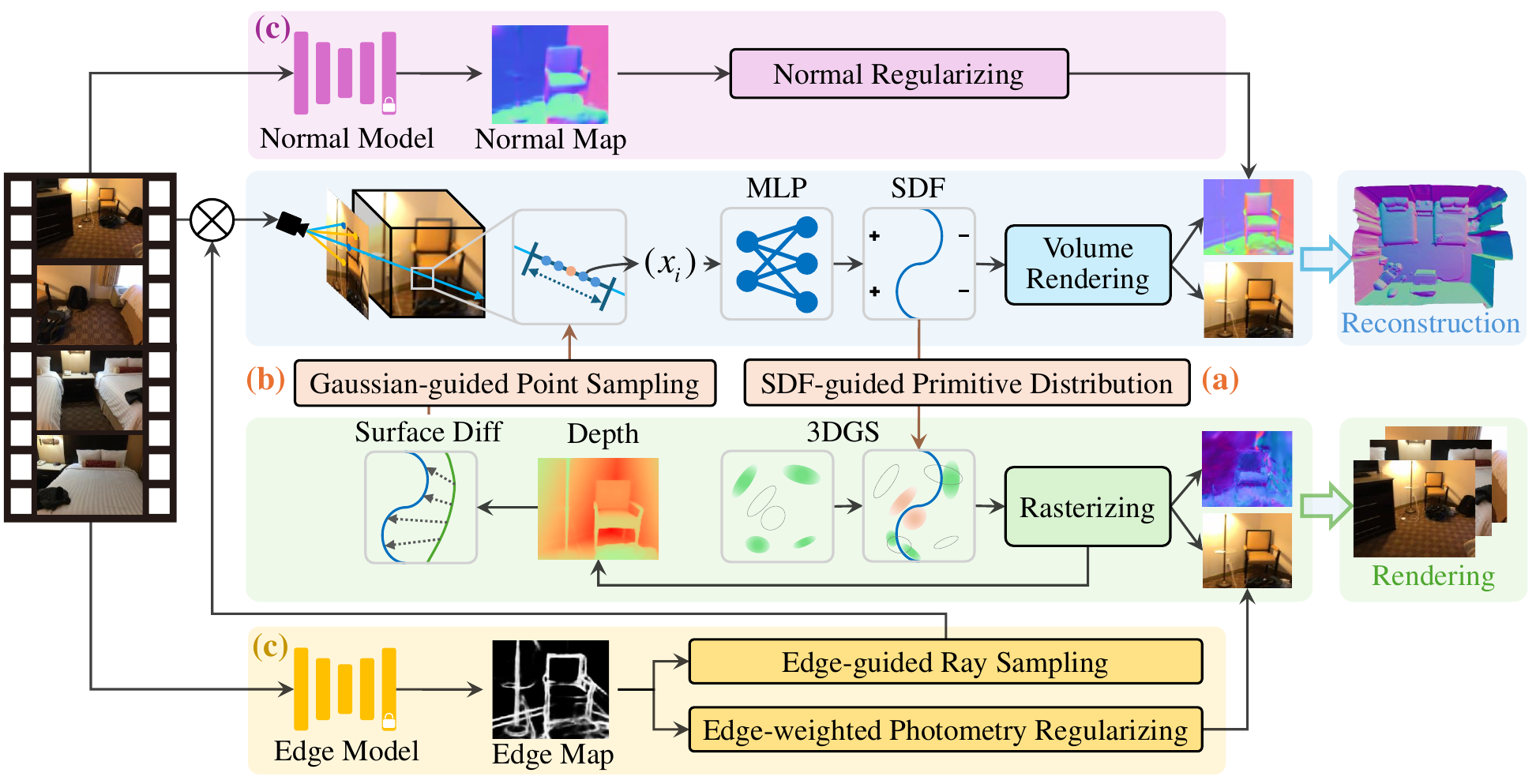}
    \vspace{-0.3cm}
    \caption{\textbf{Overview}. 
    GaussianRoom integrates neural SDF within 3DGS and forms a positive cycle improving each other. 
    (a) We employ the geometric information from the SDF to constrain the Gaussian primitives, ensuring their spatial distribution aligns with the scene surface.
    (b) We utilize rasterized depth from Gaussian to efficiently provide coarse geometry information, narrowing down the sampling range to accelerate the optimization of neural SDF.
    (c) We introduce monocular normal prior and edge prior, addressing the challenges of texture-less areas and fine structures indoors.
    }
    \label{Fig. pipeline}
    \vspace{-0.1cm}
    \vspace{-0.3cm}
\end{figure*}

Extensive experiments on ScanNet and ScanNet++ show that GaussianRoom is capable of producing high-quality reconstruction results while simultaneously maintaining the efficient rendering of 3DGS. Compared to state-of-the-art methods, our approach surpasses both rendering and reconstruction quality.
In summary, our contributions are as follows:

\begin{itemize}
  \item We propose GaussianRoom, a novel unified framework incorporating neural SDF with 3DGS. 
  An \textbf{SDF-guided primitive distribution} strategy is proposed to 
  guide the densification and pruning of Guassians. 
  At the same time, the geometry represented by Gaussians could improve the efficiency of the SDF field by piloting its point sampling.
  
  \item We design an \textbf{edge-aware regularization term} to improve the details of reconstruction, and further incorporate monocular normal priors in optimization to provide the geometric cues for textureless regions.
  
  \item Our method achieves \textbf{both high-quality surface reconstruction and rendering} for indoor scenes. Extensive experiments on various scenes show that our method achieves SOTA performance in multiple metrics.
\end{itemize}

\section{RELATED WORKS}

\vspace{-0.1cm}
\subsection{Multi-view stereo}

Feature-based Multi-View Stereo (MVS) methods \cite{barnes2009patchmatch, schonberger2016pixelwise} construct explicit representation of objects and scenes by matching image features across multiple views to estimate the 3D coordinates of pixels. Surface is then obtained by applying Poisson surface reconstruction \cite{kazhdan2013screened}. In indoor scenes, especially in large texture-less areas, these methods often struggle due to the sparsity of features. Voxel-based approaches \cite{de1999poxels} avoid the issues of poor feature matching by optimizing spatial occupancy and color within a voxel grid, but they are limited by memory usage at high resolutions, resulting in lower reconstruction quality. Learning-based multi-view stereo methods replace feature matching \cite{luo2016efficient, ummenhofer2017demon} and similar processes \cite{riegler2017octnetfusion} in feature-based methods with neural networks to directly predict depth or three-dimensional volumes from images \cite{huang2018deepmvs, yao2018mvsnet}. However, even with the use of large data during training, errors may occur in results when dealing with occlusions, complex lighting, or areas with subtle textures.

\vspace{-0.1cm}
\subsection{{Neural Radiance Field}}

Neural Radiance Fields (NeRF) \cite{mildenhall2021nerf} represents a scene as a continuous volumetric function of density and color, using neural networks to enable realistic new view synthesis. Methods such as Mip-NeRF \cite{barron2021mip} enhance rendering efficiency through optimized ray sampling techniques. Other works \cite{muller2022instant, fridovich2022plenoxels, li2023neuralangelo} accelerate training and rendering by leveraging spatial data structures, using alternative encodings, or resizing MLPs. Some works focus on improving the rendering quality through regularization terms. Depth regularization \cite{deng2022depth}, for instance, explicitly supervises ray termination to reduce unnecessary sampling time. Other approaches explore imposing smoothness constraints on rendered depth maps \cite{niemeyer2022regnerf} or employing multi-view consistency regularization in sparse view settings \cite{wang2023sparsenerf, lao2024corresnerf}. Although NeRF can produce realistic renderings for new viewpoint synthesis, the geometric quality of results directly obtained through Marching Cubes \cite{lorensen1998marching} is poor. Consequently, some research considers using alternative implicit functions, such as occupancy grids \cite{niemeyer2020differentiable, oechsle2021unisurf} and signed distance functions (SDFs) \cite{wang2022neuris, wang2021neus, li2023neuralangelo, li2024fine}, replacing NeRF’s volumetric density field. To further improve reconstruction quality, \cite{fu2022geo, zhang2022critical} propose to regularize optimization using SfM points, and \cite{guo2022neural, yu2022monosdf} leverage priors such as Manhattan world assumption and pseudo depth supervision. However, these methods tend to cause missing reconstruction while consuming a long time for optimization.

\vspace{-0.1cm}
\subsection{{3D Gaussian Splatting}}
3D Gaussian Splatting \cite{kerbl20233dgs} has recently become very popular in the field of neural rendering, providing an explicit representation of scenes and enabling novel view synthesis without the dependency on neural networks.
During training, 3DGS consumes a significant amount of GPU memory. These efforts \cite{fan2025lightgaussian,lee2024compact} aim to compress the memory footprint of Gaussian operations. GaussianPro \cite{cheng2024gaussianpro} introduces a novel Gaussian propagation strategy that guides the densification process, resulting in more compact and precise Gaussians, especially in areas with limited texture details, and similar works, such as \cite{bolanos2024gaussian,lin2024vastgaussian,hamdi2024ges}, enhance the performance of Gaussian rendering across various scenarios.
DN-Splatter \cite{turkulainen2024dnsplatter} uses depth and normal priors during the optimization process to enhance reconstruction results, achieving smoother and more geometrically accurate reconstructions. 
Similarly, SuGaR \cite{guedon2024sugar} and 2DGS \cite{huang20242d} employ Gaussian-based methods for reconstruction purposes. Our concurrent work, GSDF \cite{yu2024gsdf}, also jointly optimizes neural SDF and 3DGS. In contrast, we focus on indoor scenes and address the challenges posed by large textureless areas in 3DGS.

\begin{figure}[t]
    \centering
    \includegraphics[width=0.5\textwidth]{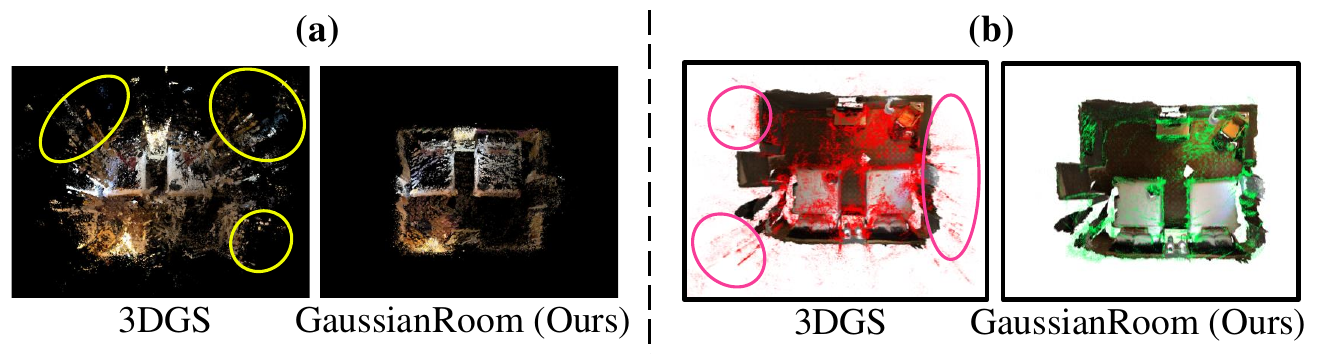}
    \vspace{-0.8cm}
    \caption{
    (a) Gaussian primitives distribution
    (b) Ground truth scene surface and Gaussian primitives distribution }
    \label{Fig.densification_ab}
    \vspace{-0.1cm}
    \vspace{-0.3cm}
\end{figure}

\section{PRELIMINARY}

\subsection{{3D Gaussian Splatting}}
3DGS \cite{kerbl20233dgs} represents the 3D scene as differentiable 3D Gaussian primitives, achieving state-of-the-art visual quality and rendering speed. Each Gaussian primitive is defined by the mean $\boldsymbol{\mu} \in \mathbb{R}^3$ and the covariance $\boldsymbol{\Sigma} \in \mathbb{R}^{3\times3}$:
\begin{equation}
    G(\boldsymbol{x})=e^{-\frac12(\boldsymbol{x}-\boldsymbol{\mu})^\mathrm{T}\boldsymbol{\Sigma}^{-1}(\boldsymbol{x}-\boldsymbol{\mu})}.
\end{equation}

To preserve the positive semi-definite character of the covariance matrix during optimization, $\boldsymbol{\Sigma}$ is further formulated as: $\boldsymbol{\Sigma}=\boldsymbol{RSS}^{\mathrm{T}}\boldsymbol{R}^{\mathrm{T}}$, where the rotation matrix $\boldsymbol{R}$ is orthogonal and scale matrix$\boldsymbol{S}$ is diagonal.
For rendering, 3D Gaussians will be projected onto the 2D image plane as 2D Gaussians following the depth-based sorting, and the color of pixel $\boldsymbol{p}$ is calculated as:
\begin{equation}
    \hat C(\boldsymbol{p})=\sum_{i\in N}\boldsymbol{c}_i\sigma_i\prod_{j=1}^{i-1}(1-\sigma_j),\quad\sigma_i=\alpha_iG_i'(\boldsymbol{p}),
\end{equation}
where $\alpha_i$ represents the opacity of the i-th 3D Gaussian, $G_i'$ is the projected 2D Gaussian, $N$ denotes the number of sorted 2D Gaussians associated with pixel $\boldsymbol{p}$, and $\boldsymbol{c}_i$ is the color of $G_i'$. Similar to \cite{cheng2024gaussianpro}, we use direction of the shortest axis as normal $n_i$ for Gaussian primitive $G_i$, and then apply $\alpha$-blending to render the normal map: $\mathcal{N}_{gs}=\sum_{i\in N}n_{i}\sigma_{i}\prod_{j=1}^{i-1}(1-\sigma_{j}).$

\subsection{Neural Implicit SDFs}

NeRF \cite{niemeyer2020differentiable} implicitly learns a 3D scene as a continuous volume density and radiance field from multi-view images. The method faces challenges in defining clear surfaces, often resulting in noisy surfaces when derived from density. In contrast, SDF offers a common way for representing geometry surfaces implicitly as a zero-level set, $\{\boldsymbol x \in \mathbb R^3 \mid f_g(\boldsymbol x)=0\}$, where $f_g(\boldsymbol{x})$ is the SDF value from an MLP $f_g(\cdot)$. Following NeuS \cite{wang2021neus}, we replace the volume density with SDF and convert the SDF value to the opacity $\alpha_i$ with a logistic function:
\begin{equation}
    \alpha_i = \max\left(\frac{\phi_s(f(\boldsymbol{x}_i))-\phi_s(f(\boldsymbol{x}_{i+1}))}{\phi_s(f(\boldsymbol{x}_i))},0\right), 
\end{equation}
where $\phi_s$ denotes a Sigmoid function. Following the volume rendering methodology, the predicted color of pixel $\boldsymbol{p}$ is calculated by accumulating weighted colors of the sample points along the ray $\boldsymbol{r}$:
\begin{equation}
    \Hat C(\boldsymbol{p})=\sum_{i=1}^{N}T_i\alpha_i\boldsymbol{c_i},T_i=\exp\left(-\sum_{j=1}^{i-1}\alpha_j\delta_j\right), 
\end{equation}
where $T_i$ is the cumulative transmittance and $N$ is the number of sample points along the ray $\boldsymbol{r}$. Similarly, the normal can be rendered as $\hat{\mathcal{N}}(\boldsymbol{p})=\sum_{i=1}^N T_i\alpha_i\hat{\boldsymbol n}_i$, where $\hat{\boldsymbol n}_i=\nabla f_n(t_i)$ denotes the derivative of SDF at point $t_i$, which can be calculated by PyTorch’s automatic derivation.

\section{METHODOLOGY} \label{METHODOLOGY}

Given multi-view posed images and point clouds obtained from the Structure from Motion (SfM) algorithm, our objective is to enable 3DGS to reconstruct indoor scene geometry accurately while preserving its rendering quality and efficiency. 
To this end, we first incorporate an implicit SDF field within 3DGS and design a mutual learning strategy to realize high-quality reconstruction and rendering (Sec. \ref{sec: Mutual Learning}). Furthermore, we present the monocular geometric cues, i.e. normal and edge prior, to improve the reconstruction of the planes and fine details in indoor scenes respectively (Sec. \ref{sec: Leveraging Monocular Cues}). Finally, we discuss the loss functions and the overall optimization process (Sec. \ref{sec: Loss Functions}). An overview of our framework is provided in Fig. \ref{Fig. pipeline}.

\vspace{-0.1cm}
\subsection{Mutual Learning of 3D Gaussian and Neural SDF} \label{sec: Mutual Learning}


In this section, we propose to model an indoor scene using 3D Gaussians and neural SDF with a mutual learning strategy. For 3DGS optimization, we utilize SDF to guide the distribution of Gaussian primitives, which densifies the Gaussians around the surface and significantly reduces the floaters in non-surface space. Meanwhile, we introduce a Gaussian-guided sampling methodology to pilot point sampling for neural SDF, which improves the training efficiency.

\begin{figure}[t]
    \centering
    \includegraphics[width=0.5\textwidth]{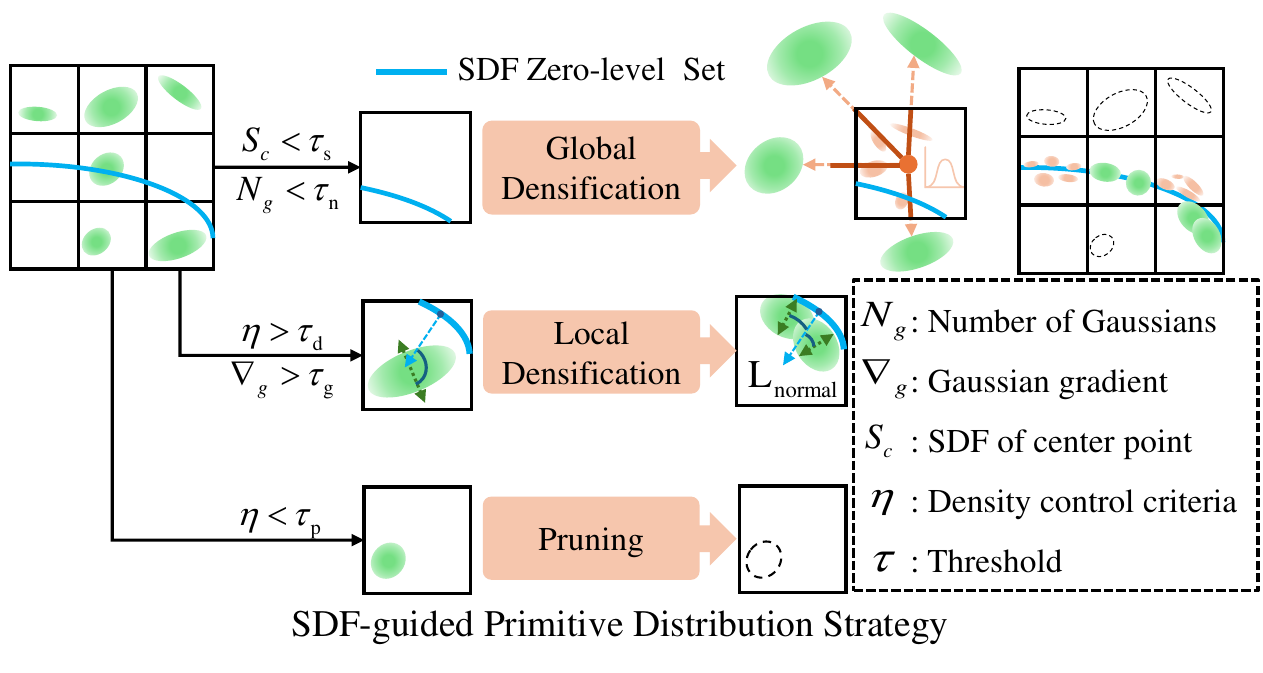}
    \vspace{-0.8cm}
    \caption{The red Gaussian points represent new Gaussians generated by the SDF-guided Global Densification strategy, while the green Gaussian points indicate those adjusted through the SDF-guided Densification and Pruning process. }
    \label{Fig.densification_c}
    \vspace{-0.2cm}
    \vspace{-0.3cm}
\end{figure}

\noindent\textbf{SDF-guided Primitive Distribution for 3D Gaussian}
As illustrated in Fig. \ref{Fig.densification_ab}, due to the lack of underlying geometry constrain, Gaussian primitives become disorganized during the optimization in indoor scenes, resulting in randomly scattered floaters in the non-surface areas. 
Specifically, the guidance of the SDF includes two aspects, one is to deploy Gaussian primitives to achieve global densification in spatial locations with low SDF values that lack Gaussian primitives, and the other is to guide the densification and pruning of existing Gaussian primitives.
\textit{SDF-guided Global Densification.} 
Since the original 3DGS method densifies Gaussians based on existing ones, it struggles to generate new Gaussian primitives in regions lacking initial Gaussians. This issue becomes more pronounced in indoor scenes, where texture-less areas often exhibit a scarcity of Gaussians. To address this issue, we develop a global densification strategy that takes advantage of the geometric information of the entire scene provided by the neural SDF.

As depicted by the Global Densification strategy in Fig. \ref{Fig.densification_c}, we partition the scene space into $\mathrm{N}^3$ cubic grids and calculate the SDF value at the center of each grid. If the value falls below the threshold ($S_c < \tau_s$), it indicates that the grid is in proximity to the scene surface. Subsequently, we enumerate the existing Gaussian primitives within each grid. In cases where the number of Gaussian primitives is insufficient ($N_g < \tau_n$), we select the $K$ Gaussian neighbors of the grid’s center point and generate $K$ new Gaussian primitives within the grid. The initial attributes of these newly generated Gaussian primitives are sampled from a normal distribution defined by the mean and variance of the $K$ neighboring Gaussians.

\textit{SDF-guided Densification and Pruning.}
For regions where a sufficient number of Gaussian primitives already exist, we employ an enhanced version of the Densification and Pruning strategy, which integrates SDF geometric information.
For each Gaussian primitive at position $\boldsymbol{x}$, its SDF value is given by: $S = f_g(\boldsymbol{x})$. The criteria for determining whether a Gaussian primitive should be densified or pruned can be expressed as follows:
\begin{equation} \label{eq: eta}
    \eta= \exp(- \frac{S^2}{\lambda_{\sigma}\sigma^2}),
\end{equation}
where $\sigma$ denotes the opacity of the Gaussian primitive, and $\lambda_{\sigma}$ is its coefficient. When $\eta$ is small, it signifies that the Gaussian is either far from the SDF zero-level set or possesses low opacity. In such instances, if $\eta < \tau_p$, the Gaussian primitive will be pruned. Conversely, when $\eta > \tau_d$ and the gradient of the Gaussian satisfies $\nabla_g > \tau_g$, the Gaussian primitive will be densified. GSDF \cite{yu2024gsdf} introduces a similar method, but our defined $\eta$ in Eq. \ref{eq: eta} integrates both the SDF value and the Gaussian's opacity while serving as separate criteria independent of the gradient of Gaussian, thus avoiding the trade-off between densifying Gaussians in high-gradient regions and removing floaters. In other words, our methods will not mistakenly densify Gaussians detached from the SDF surface.

\noindent\textbf{Gaussian-guided Point Sampling for Neural SDF}
To reconstruct the accurate geometry surface efficiently, it is advisable to sample as many points around the true surface as possible. 
Previous works \cite{sun2022neural, li2023edge} use the predicted SDF values to pilot sampling along the ray. However, this approach is time-consuming and suffers from the chicken and egg problem. Instead, we employ the 3D Gaussians as coarse geometry guidance for point sampling similar to \cite{yu2024gsdf}. Specifically, we leverage the rasterized depth maps from the 3D Gaussians to narrow down the ray sampling range of the neural SDF field. 
The rendered depth value $D$ of pixel $\boldsymbol{p}$ rendered is defined as follows:
\begin{equation}
    D(\boldsymbol p)=\sum_{i\in N}d_i\sigma_i\prod_{j=1}^{i-1}(1-\sigma_j),
\end{equation}
where $N$ is the number of 3D Gaussians encountered by the ray $\boldsymbol r$ corresponding to $\boldsymbol p$, $d_i$ represents the depth of the i-th 3D Gaussian and $\sigma_i$ demotes the opacity of projected Gaussian primitive. 

Once the rendered depth $D(\boldsymbol p)$ has been obtained, we define the sampling range as follows: $\boldsymbol{r} = [\boldsymbol{o} + (D(\boldsymbol{p}) - \gamma\left| S \right|) \cdot \boldsymbol{v}, \boldsymbol{o} + (D(\boldsymbol{p}) + \gamma\left| S \right|) \cdot \boldsymbol{v}]$, where $\boldsymbol{o}$ and $\boldsymbol{v}$ respectively denotes the camera center and view direction of pixel $\boldsymbol{p}$, $S$ represents the corresponding SDF value, and $\gamma$ is a hyperparameter indicating the length of the sampling range. 

\vspace{-0.1cm}
\subsection{Monocular Cues constrained Optimization} \label{sec: Leveraging Monocular Cues}
It is important to note that indoor scenes encompass not only large texture-less areas such as walls and floors but also detailed regions. However, in the original 3DGS and neural SDF, the optimization only relies on image reconstruction loss without incorporating any geometric constraints, resulting in noisy and blurry results. In this section, we incorporate monocular geometric cues into our designed mutual learning pipeline, including edge prior and normal prior, which are used to constrain details and flat areas, respectively.

\noindent\textbf{Edge-guided Details Optimization}
Noticing that detailed regions account for a small proportion of indoor scenes compared with texture-less flat areas like floors and ground, leading to relative ignorance of the details. Based on the observation that detailed regions mostly have sharp shapes or rich textures with obvious edge information, we propose a novel edge-guided optimization strategy. We use a pre-trained edge detection network \cite{su2021pixel} to generate edge maps. 

\textit{Edge-aware Neural SDF Optimization}. For neural SDF, we design an edge-guided ray sampling strategy, which performs explicit sampling at detailed areas. Specifically, we define an image-variant weight for ray sampling according to the ratio of edge pixels: 
\begin{equation}
    \omega_i = \delta \cdot N_{edge}^i/(H\times W),
\end{equation}
where $N_{edge}^i$ is the number of edge pixels for image $I_i$, $H\times W$ is the total number of pixels in image $I_i$, and $\delta$ is a hyperparameter indicating the importance of edge. For the $q$ sampled training rays of image $I_i$, $\omega_i*q$ rays are sampled from the edge region set and $(1-\omega_i)*q$ rays are sampled from a random set. The hybrid sampling method ensures that sharp boundary information can be sampled in each iteration adaptively. 

\textit{Edge-aware 3DGS Optimization}. Considering regions with high-frequency information that are relatively difficult to recover, we design a simple yet effective regularizing strategy to improve the rendering results of details. Our method is developed based on the insight that pixels in details and pixels in flat areas have the same influence on the result in 3DGS, which is inefficient in indoor scenes. Thus, we utilize edge priors as weights to regularize the photometric consistency. For each pixel $\boldsymbol{p}$ of image $I_i$, the corresponding training weight of the photometric loss is defined as follows:
\begin{equation}
    w_{\boldsymbol{p}} = 2\phi(e_{\boldsymbol{p}}),
\end{equation}
where $e_{\boldsymbol{p}}$ represents corresponding edge map value for pixel $\boldsymbol p$ and $\phi(\cdot)$ denotes a Sigmoid function. We utilize the mapping function to confine the loss weight within the interval $\left[1, 2\right]$ and increase the importance of edge areas in the optimization, which is beneficial to recover fine details. 

\noindent\textbf{Normal-guided Geometry Constrain}
Based on the observation that texture-less areas usually exhibit well-defined planarity, we utilize normal information to provide geometry constraints. 
Specifically, we use \cite{bae2021estimating} to get monocular normal priors, and inspired from \cite{yu2022monosdf, cheng2024gaussianpro}, we utilize the prior to simultaneously constrain the rendered normal maps of neural SDF and 3DGS. 

\subsection{Loss Functions} \label{sec: Loss Functions}
The Gaussians is supervised by rendering losses $\mathcal{L}_{c}$, $\mathcal{L}_{D-SSIM}$ and normal loss $\mathcal{L}_{normal}$: 
\begin{equation}
    \mathcal{L}_{gs}=\lambda_{1}\mathcal{L}_{c}+(1 - \lambda_{1})\mathcal{L}_{D-SSIM}+\lambda_{2}\mathcal{L}_{normal},
\end{equation}
where the rendering loss $\mathcal{L}_{c}$ is defined as:
\begin{equation}
    \mathcal{L}_{c}=\frac{1}{q}\sum_k \|C_k-\hat{C}_k\|_1 \cdot w_k,
\end{equation}
where $C_k$ and $\hat{C}_k$ are ground truth and rendered colors respectively, $w_k$ is the weighted term from the edge map mentioned in Sec. \ref{sec: Leveraging Monocular Cues}. The normal loss $\mathcal{L}_{normal}$ is defined as:
\begin{equation}
    \mathcal{L}_{normal}=\frac{1}{q}\sum_k\|\mathcal{N}_k-\hat{\mathcal{N}}_k\|_1,
\end{equation}
where $\mathcal{N}_k$ and $\hat{\mathcal{N}}_k$ denote the predicted monocular normals and rendered normals respectively.

The neural SDF is supervised by rendering loss $\mathcal{L}_{c}$, normal loss $\mathcal{L}_{normal}$ and Eikonal loss $\mathcal{L}_{eik}$:
\begin{equation}
    \mathcal{L}_{sdf}=\mathcal{L}_{c}+\mathcal{L}_{normal}+\lambda_{eik}\mathcal{L}_{eik},
\end{equation}
where the Eikonal loss is to regularize the SDF following \cite{gropp2020implicit}. 
Our total loss is defined as:
\begin{equation}
    \mathcal{L}=\mathcal{L}_{gs}+\mathcal{L}_{sdf}. 
\end{equation}

\begin{figure}[!t]
    \centering
    \includegraphics[width=0.5\textwidth]{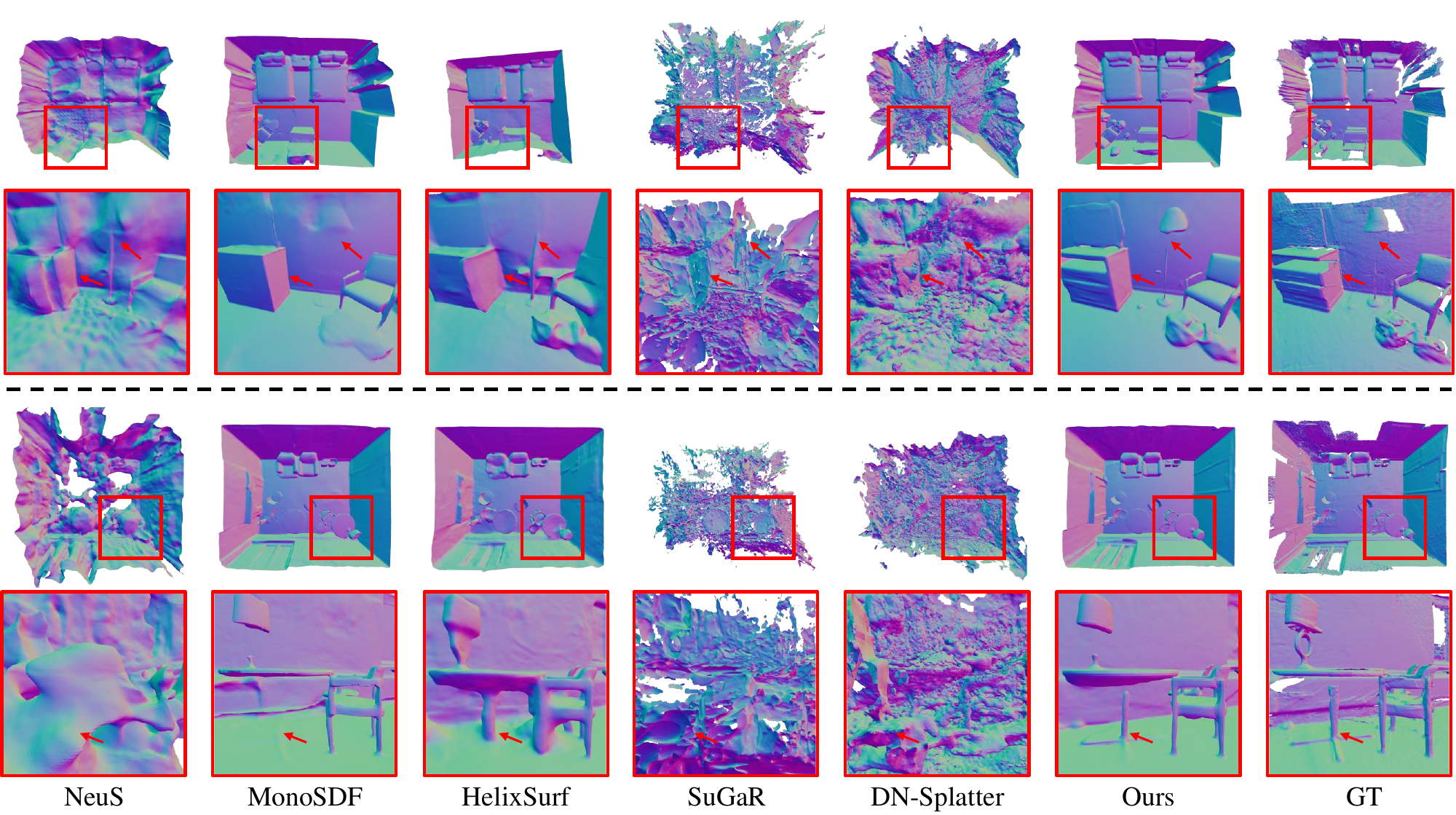}
    \vspace{-0.5cm}
    \caption{\textbf{Qualitative reconstruction comparisons}. For each indoor scene, the first row is the top view of the whole room, and the second row is the details of the masked region. The reconstruction results of GaussianRoom visually have better scene integrity than other methods, especially in details.}
    \label{Fig. vis-comp}
    \vspace{-0.3cm}
\end{figure}

\vspace{-0.1cm}
\section{Experiment}\label{sec:Experiment}

\subsection{Experimental Setup} \label{sec: Experimental Setup}

\textbf{Dataset} 
We evaluate the performance of our approach on reconstruction and rendering quality using 10 real-world indoor scenes from publicly available datasets: 8 scenes from ScanNet(V2) \cite{dai2017scannet} and 2 scenes from ScanNet++ \cite{yeshwanth2023scannet++}.

\noindent\textbf{Implementation Details} 
We build our code based on 3DGS \cite{kerbl20233dgs} and NeuS \cite{wang2021neus}. During the optimization, we divide the training into three stages. We first pre-train 3DGS with 15k iterations and then co-optimize 80k iterations, of which the 3DGS and the neural SDF don't instruct each other for the first 6k to ensure that they each learn rough information about the scene before mutual learning. 
During the co-optimization stage of 3DGS and neural SDF, we implement the SDF-guided Densification and Pruning strategy every 100 iterations and the SDF-guided Global Densification strategy every 2000 iterations, rather than employing the original Gaussian densification and pruning strategy, with other hyper-parameters remaining mostly consistent with 3DGS. As for the neural SDF, We sample 1024 rays per batch and 64+64 points on each ray. After training, we extract a mesh from the SDF by the Marching Cube algorithm \cite{lorensen1998marching} with the volume size of $512^{3}$. For geometry comparison, we utilize Poisson reconstruction \cite{kazhdan2006poisson} to get the mesh for COLMAP \cite{schonberger2016structure} and some Gaussian-based methods. 

\noindent\textbf{Baselines} 
We evaluate our method against SOTA reconstruction and rendering approaches respectively. In terms of reconstruction, we compare with COLMAP \cite{schonberger2016structure}, NeRF \cite{mildenhall2021nerf}, NeuS \cite{wang2021neus}, MonoSDF \cite{yu2022monosdf}, HelixSurf \cite{liang2023helixsurf}, 3DGS \cite{kerbl20233dgs}, GaussianPro \cite{cheng2024gaussianpro}, SuGaR \cite{guedon2024sugar} and DN-Splatter \cite{turkulainen2024dnsplatter}. For rendering, we compare with Gaussian-based methods due to their impressive rendering quality, including 3DGS \cite{kerbl20233dgs}, SuGaR \cite{guedon2024sugar}, GaussianPro \cite{cheng2024gaussianpro} and DN-Splatter \cite{turkulainen2024dnsplatter}. 
For DN-Splatter \cite{schonberger2016structure}, we follow their optional experimental settings of using monocular depths instead of real sensor depths as supervision and retrain DN-Splatter, ensuring the comparison is fair. 


\noindent\textbf{Metrics} 
We follow the evaluation protocol from \cite{wang2022neuris, yu2022monosdf} and report Accuracy, Completion, Precision, Recall, and F-score with a threshold of 5cm for 3D geometry evaluation. For rendering evaluation, we follow standard practice and report SSIM, PSNR, and LPIPS metrics.

\vspace{-0.1cm}
\subsection{Reconstruction Evaluation}
Tab. \ref{tab:table1} shows GaussianRoom outperforms both Gaussian-based and NeRF-based methods in geometry metrics on the ScanNet and ScanNet++ datasets. 
As illustrated in Fig. \ref{Fig. vis-comp}, Gaussian-based scene reconstruction methods, such as SuGaR \cite{guedon2024sugar} and DN-Splatter \cite{turkulainen2024dnsplatter}, are impacted by the geometric disorder of Gaussians, leading to uneven Poisson reconstructions. 
In contrast, our methods can achieve smooth and continuous results due to the continuity of neural SDF. Compared with NeRF-based methods, GaussianRoom achieves more complete and detailed reconstruction while greatly shortening training time, due to integrated 3DGS encouraging more efficient point sampling near the surface and edge prior offers more attention to detailed regions. For example, MonoSDF requires about 18-hour optimization, whereas GaussianRoom achieves better results within 5-hour training. 

\begin{table}[!h]
    \caption{\textbf{Quantitative reconstruction comparison} on ScanNet / ScanNet++. Average results are reported over 8 and 2 scenes, respectively, with the best results in \textbf{bold}.}
    \vspace{-0.2cm}
    \label{tab:table1}
    \centering
    \scalebox{0.65}
    {
    \begin{tabular}{l|cccc>{\columncolor[gray]{0.902}}c}
    \hline
    Method      & Accuracy$\downarrow$     & Completion$\downarrow$   & Precision$\uparrow$    & Recall$\uparrow$       & \textbf{F-score}$\uparrow$      \\ \hline
    COLMAP \cite{schonberger2016structure}      & 0.062 / 0.091 & 0.090 / 0.093 & 0.640 / 0.519 & 0.569 / 0.520 & 0.600 / 0.517 \\ \hline
    NeRF \cite{mildenhall2021nerf}        & 0.160 / 0.135 & 0.065 / 0.082 & 0.378 / 0.421 & 0.576 / 0.569 & 0.454 / 0.484 \\
    NeuS \cite{wang2021neus}        & 0.105 / 0.163 & 0.124 / 0.196 & 0.448 / 0.316 & 0.378 / 0.265 & 0.409 / 0.288 \\
    MonoSDF \cite{yu2022monosdf}     & 0.048 / 0.039 & 0.068 / 0.043 & 0.673 / 0.816 & 0.558 / 0.840 & 0.609 / 0.827 \\
    HelixSurf \cite{liang2023helixsurf}   & 0.063 / \ \ \ --\ \ \  & 0.134 / \ \ \ --\ \ \     & 0.657 / \ \ \ --\ \ \     & 0.504 / \ \ \ --\ \ \     & 0.567 / \ \ \ --\ \ \    \\ \hline
    3DGS \cite{kerbl20233dgs} & 0.338 / 0.113 & 0.406 / 0.790 & 0.129 / 0.445 & 0.067 / 0.103 & 0.085 / 0.163 \\
    GaussianPro \cite{cheng2024gaussianpro} & 0.313 / 0.141 & 0.394 / 1.283 & 0.112 / 0.353 & 0.075 / 0.081 & 0.088 / 0.129 \\
    SuGaR \cite{guedon2024sugar}       & 0.167 / 0.129 & 0.148 / 0.121 & 0.361 / 0.435 & 0.373 / 0.444 & 0.366 / 0.439 \\ 
    DN-Splatter \cite{turkulainen2024dnsplatter} & 0.212 / 0.294 & 0.210 / 0.276 & 0.153 / 0.108  & 0.182 / 0.108 & 0.166 / 0.107 \\ 
    2DGS \cite{huang20242d} & 0.167 / \ \ \ --\ \ \ & 0.152 / \ \ \ --\ \ \  & 0.311 / \ \ \ --\ \ \  & 0.341 / \ \ \ --\ \ \  & 0.324 / \ \ \ --\ \ \  \\ \hline
    Ours        & \textbf{0.047} / \textbf{0.035} & \textbf{0.043} / \textbf{0.037} & \textbf{0.800} / \textbf{0.894} & \textbf{0.739} / \textbf{0.852} & \textbf{0.768} / \textbf{0.872} \\ \hline
    \end{tabular}
    }

\end{table}
\vspace{-0.5cm}
\begin{table}[!h]
\caption{\textbf{Quantitative rendering comparison} with existing methods on ScanNet and ScanNet++ datasets. We report the average results over 8 and 2 scenes respectively.}
\label{tab:2D_compare_quality}
\centering

\scalebox{0.7}
{\begin{tabular}{@{}l|ccc|ccc@{}}
\toprule
            & \multicolumn{3}{c|}{ScanNet}                                                                                                      & \multicolumn{3}{c}{ScanNet++}                                                                                   \\
Method      & \multicolumn{1}{l}{SSIM$\uparrow$} & \multicolumn{1}{l}{PSNR$\uparrow$} & \multicolumn{1}{l|}{LPIPS$\downarrow$} & \multicolumn{1}{l}{SSIM$\uparrow$} & \multicolumn{1}{l}{PSNR$\uparrow$} & \multicolumn{1}{l}{LPIPS$\downarrow$} \\ \midrule
3DGS  \cite{kerbl20233dgs}       & \cellcolor[HTML]{FFFFB2}0.731      & 22.133                                              & \cellcolor[HTML]{FFFFB2}0.387          & \cellcolor[HTML]{FFD8B2}0.843      & \cellcolor[HTML]{FFD8B2}21.816     & \cellcolor[HTML]{FFD8B2}0.294         \\
SuGaR \cite{guedon2024sugar}       & \cellcolor[HTML]{FFD8B2}0.737      & \cellcolor[HTML]{FFFFB2}22.290                      & \cellcolor[HTML]{FFD8B2}0.382          & \cellcolor[HTML]{FFFFB2}0.831      & 20.611     & 0.318                                 \\
GaussianPro \cite{cheng2024gaussianpro} \ \ \ \ \ \ \ \ \ \ \ \ \ \ \ \ \ \ \ \ \ \ \ \ \  & 0.721                              & \cellcolor[HTML]{FFD8B2}22.676                      & 0.395                                  & \cellcolor[HTML]{FFFFB2}0.831      & \cellcolor[HTML]{FFFFB2}21.285                             & 0.320                                 \\
DN-Splatter \cite{turkulainen2024dnsplatter} & 0.639                              & 21.621                                              & \cellcolor[HTML]{FFB2B2}0.312          & 0.826                              & 20.445                             & \cellcolor[HTML]{FFB2B2}0.268         \\
Ours        & \cellcolor[HTML]{FFB2B2}0.758      & \cellcolor[HTML]{FFB2B2}23.601                      & 0.391                                  & \cellcolor[HTML]{FFB2B2}0.844      & \cellcolor[HTML]{FFB2B2}22.001     & \cellcolor[HTML]{FFFFB2}0.296         \\ \bottomrule
\end{tabular}

}

\vspace{-0.3cm}
\end{table}

\begin{figure}[h]
    \centering
    \includegraphics[width=0.5\textwidth]{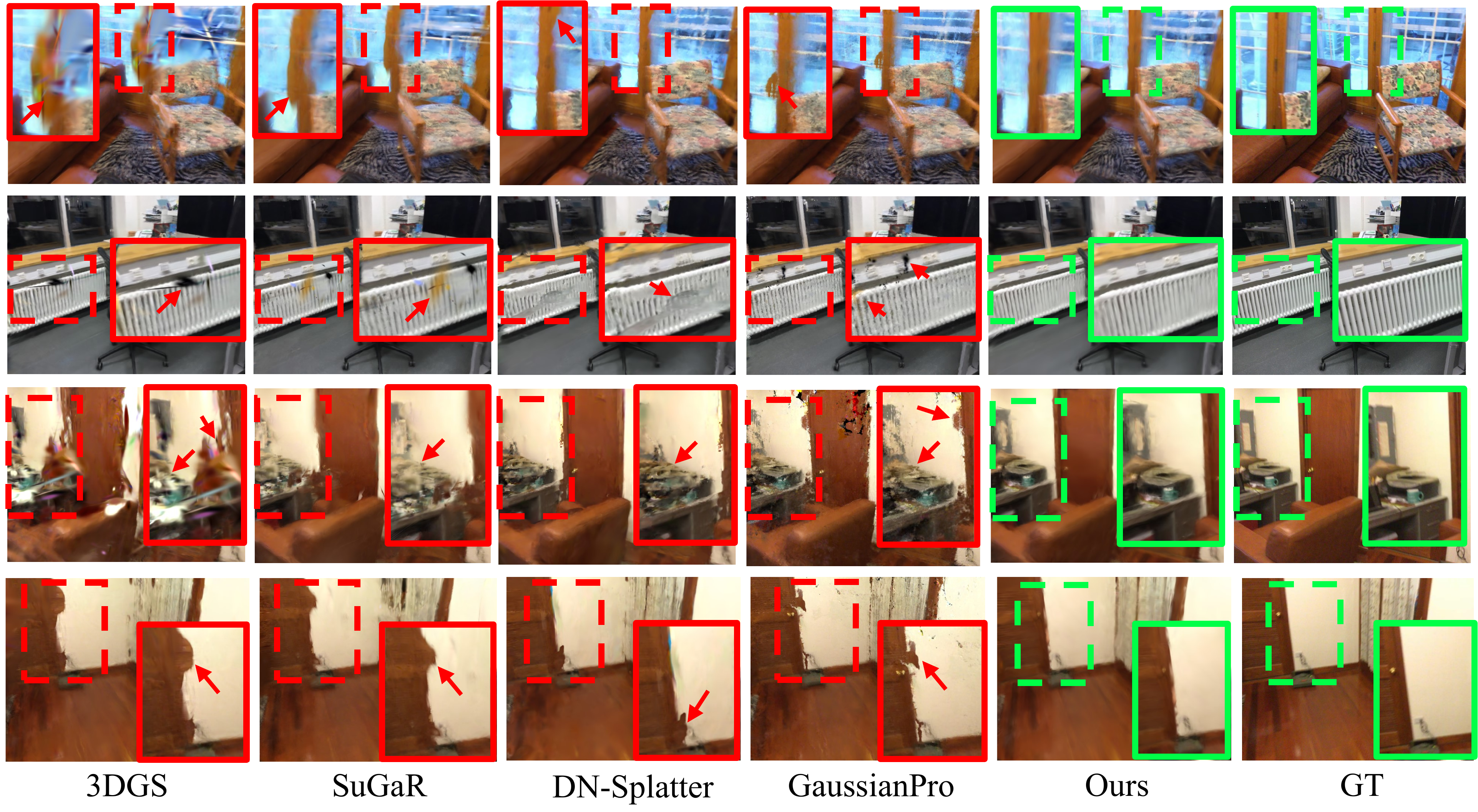}
    \vspace{-0.5cm}
    \caption{\textbf{Qualitative rendering comparisons}. As shown from the above-highlighted patches, the rendering results of GaussianRoom outperform other GS-based methods, including texture-less regions and details.}
   \label{Fig.render_compare}
   \vspace{-0.4cm}
\end{figure}

\subsection{Rendering Evaluation}
According to Tab. \ref{tab:2D_compare_quality}, our method demonstrates higher rendering metrics on both the ScanNet and ScanNet++ datasets compared to 3DGS \cite{kerbl20233dgs}, SuGaR \cite{guedon2024sugar}, GaussianPro \cite{cheng2024gaussianpro}, and DN-Splatter \cite{turkulainen2024dnsplatter}.
The substantial improvements in the PSNR metric indicate that our rendered images exhibit reduced distortion, credited to our designed SDF-guided Primitive Distribution Strategy, which effectively prunes floating Gaussian primitives in non-surface space and deploys Gaussians near the surface that lacks initial Gaussians, as demonstrated in Fig. \ref{Fig.densification_c}. 
As shown in Fig. \ref{Fig.render_compare}, our method achieves sharp details and improved renderings in both rich-texture and low-texture areas by leveraging geometric information in the neural SDF to align Gaussian primitives with the scene surface.

\vspace{-0.1cm}
\subsection{Ablation Study}
In this section, we conduct ablation experiments to verify the superiority of mutual learning of 3DGS and SDF fields and the effectiveness of monocular cues. As shown in Tab. \ref{tab:new_ablation}, training neural SDF is time-consuming and its rendering quality and speed are inferior compared to 3DGS. In contrast, 3DGS struggles to learn geometry accurately, resulting in poor reconstruction quality. Compared to them, GaussianRoom not only avoids their weaknesses but also brings further improvements in both reconstruction and rendering quality. GaussianRoom requires only half the training time of SDF while achieving improved reconstruction results. Additionally, its rendering quality surpasses that of 3DGS with monocular cues, while maintaining a rendering speed of over 170 fps.


\begin{table}[]
\caption{\textbf{Ablation study of mutual optimization}. The ablation experiments were conducted on 8 scenes from the ScanNet dataset, with the best metrics highlighted in \textbf{bold}.}
\vspace{-0.2cm}
\label{tab:new_ablation}

\resizebox{0.5\textwidth}{!}{
\centering
\begin{tabular}{@{}c|c|ccc|c|c@{}}
\toprule
                    & 3D Reconstruction & \multicolumn{3}{c|}{Novel View Systhesis} & Rendering speed &               \\
Method              & F-score $\uparrow$  & SSIM $\uparrow$       & PSNR $\uparrow$      & LPIPS $\downarrow$    & fps $\uparrow$           & Training time $\downarrow$ \\ \midrule
GaussianRoom(full)        & \textbf{0.768}    & \textbf{0.758}      & {23.601}     & \textbf{0.391}     & \textbf{170+}          & $\sim$4.5h    \\
SDF + MonoCues      & 0.719    & 0.714      & 22.634     & 0.458     & $<$1            & $\sim$9h      \\
3DGS + MonoCues     & 0.163    & 0.726      & 22.092     & 0.396     & \textbf{170+}            & \textbf{$\sim$1h}      
\\
w/o normal prior     & 0.388    & 0.755      & \textbf{23.625}     & 0.395     & \textbf{170+}            & $\sim$4.5h
\\
w/o edge prior    & 0.750    & 0.754      & 23.35     & 0.397     & \textbf{170+}            & $\sim$4.5h
\\ \bottomrule
\end{tabular}
}
\vspace{-0.7cm}
\end{table}
Ablation studies in Tab. \ref{tab:new_ablation} also show that the normal prior and edge prior improve 3D reconstruction quality, with each primarily functioning in low-texture regions and detailed regions, respectively. Given the large low-texture areas typical in indoor reconstruction tasks, the supervision provided by the normal prior proves to be highly effective. However, similar to DN-Splatter \cite{turkulainen2024dnsplatter}, utilizing the normal prior as a regularization term to constrain the orientation of Gaussians results in a decrease in the PSNR metric. Despite this, our full model excels in both SSIM and LPIPS metrics. 

\vspace{-0.1cm}
\section{CONCLUSIONS}
\vspace{-0.1cm}
\label{sec:Conclusion}
In conclusion, we present a novel unified framework that combines neural SDF with 3DGS. By incorporating a learnable neural SDF field with the SDF-guided primitive distribution strategy, our method overcomes the limitations of 3DGS in reconstructing indoor scenes with textureless areas. Meanwhile, the integration of Gaussians improves the efficiency of the neural SDF, while regularization with normal and edge priors enhances geometry details. Extensive experiments demonstrate the state-of-the-art performance of our method in surface reconstruction and novel view synthesis on ScanNet and ScanNet++ datasets.

\section*{ACKNOWLEDGMENT}
This research was funded by Shenzhen Major Science and Technology Project (KJZD20230923115503007)




\newpage
\bibliographystyle{IEEEtrans}  
\bibliography{IEEEfull}     

\end{document}